\documentclass[10pt,twocolumn,letterpaper]{article}

\usepackage{wacv}              

\usepackage{graphicx}
\usepackage{amsmath}
\usepackage{amssymb}
\usepackage{booktabs}

\usepackage[pagebackref,breaklinks,colorlinks]{hyperref}

\usepackage[capitalize]{cleveref}
\crefname{section}{Sec.}{Secs.}
\Crefname{section}{Section}{Sections}
\Crefname{table}{Table}{Tables}
\crefname{table}{Tab.}{Tabs.}

\usepackage{color,soul}
\usepackage{float}
\usepackage[table,xcdraw]{xcolor}
\usepackage{multirow}
\usepackage{hhline}
\usepackage{arydshln}
\newcommand*{\proposed}{\textbf{Sli2Vol+}\@\xspace}
\usepackage{rotating}

\def\wacvPaperID{247} 
\def\confName{WACV}
\def\confYear{2025}

\usepackage[accsupp]{axessibility}

\begin{document}

\title{Sli2Vol+: Segmenting 3D Medical Images Based on an Object Estimation Guided \\ Correspondence Flow Network\thanks{$\star$ indicates equal contribution. This research was supported in part by NSF grants: IIS-1955395, IIS-2101696, OAC-2104158, IIS-2401144, NIH grants: 1R01HL177814-01, R01EB004640, 2R56EB004640-16, and NIH/NIBIB R56 EB004640.}}

\author{Delin An$^{\star}$\\
University of Notre Dame\\
{\tt\small dan3@nd.edu}
\and
Pengfei Gu$^{\star}$\\
The University of Texas Rio Grande Valley\\
{\tt\small pengfei.gu01@utrgv.edu}
\and
Milan Sonka\\
University of Iowa\\
{\tt\small milan-sonka@uiowa.edu}
\and 
Chaoli Wang\\
University of Notre Dame\\
{\tt\small chaoli.wang@nd.edu}
\and
Danny Z. Chen\\
University of Notre Dame\\
{\tt\small dchen@nd.edu}
}
\maketitle

\begin{abstract}
Deep learning (DL) methods have shown remarkable successes in medical image segmentation, 
often using large amounts of annotated data for model training.
%
However, acquiring a large number of diverse labeled 3D medical image datasets is highly difficult and expensive.
Recently, mask propagation DL methods were developed to reduce the annotation burden on 3D medical images.
For example, Sli2Vol~\cite{yeung2021sli2vol} proposed a 
self-supervised framework (SSF)
to learn correspondences by matching neighboring slices via slice reconstruction in the training stage;
the learned correspondences were then used to propagate a labeled slice to other slices in the test stage.
But, these methods are still prone to error accumulation
due to the inter-slice propagation of reconstruction errors. Also, they do not handle discontinuities well, 
which can occur between consecutive slices in 3D images, as they emphasize exploiting object continuity.
To address these challenges, in this work, we propose a new 
SSF, called \proposed, {for segmenting any anatomical structures in 3D medical images using only a single annotated slice per training and testing volume.}
Specifically, in the training stage, we first propagate an annotated 2D slice of a training volume to the other slices, generating pseudo-labels (PLs). 
Then, we develop a novel Object Estimation Guided Correspondence Flow Network 
to learn reliable correspondences between consecutive slices and corresponding PLs in a self-supervised manner.
In the test stage, such correspondences are utilized to propagate a single annotated slice to the other slices of a test volume.
%
We demonstrate the effectiveness of our method on various medical image segmentation tasks with different datasets, showing better generalizability across different organs, modalities, and modals. 
Code is available at \url{https://github.com/adlsn/Sli2Volplus} 
\end{abstract}


\section{Introduction}
\label{sec:intro}
Image segmentation is a critical task in medical image analysis, providing anatomical structure information essential for disease diagnosis and treatment planning~\cite{celebi2007methodological,pace2015interactive}. 
Known deep learning (DL) methods have achieved state-of-the-art (SOTA) performance in many medical image segmentation tasks, including convolutional neural network (CNN)-based methods~\cite{ronneberger2015u,oktay2018attention,zhou2018unet++,isensee2021nnu,gu2021k}, Transformer-based methods~\cite{karimi2021convolution,cao2023swin}, and hybrid approaches~\cite{hatamizadeh2022unetr,zhou2021nnformer,shaker2022unetr++,gu2022convformer}.
However, these DL methods require abundant labeled training data to attain satisfactory performance. Labeling large amounts of medical image data, especially for 3D images, is highly difficult and expensive as this process requires domain-specific expertise, and pixel/voxel-wise annotations can be very labor-intensive and time-consuming.

Four main methods have been proposed to address the data-annotation burden. 
The first type explores the potential of non-annotated data through semi-supervised~\cite{radosavovic2018data,zhang2017deepvovic2018data,zhou2018semi} and self-supervised~\cite{tao2020revisiting,zhang2022keep,zhang2023point,zhou2019models} methods to reduce the demand for labeled data.
The second type leverages the segment anything model (SAM)~\cite{kirillov2023segment} for medical image segmentation (e.g., \cite{deng2023segment,hu2023sam,he2023accuracy,roy2023sam,mohapatra2023sam,ma2023segment,mazurowski2023segment,zhou2023can}). However, the applicability of these methods to medical image segmentation remains limited due to the significant differences between natural images and medical images.
The third type is weakly-supervised methods, such as image-level~\cite{chang2020weakly,huang2018weakly,kolesnikov2016seed}, {patch-level~\cite{dang2022vessel,xu2014weakly,xu2019camel},} bounding box~\cite{song2019box,papandreou2015weakly,khoreva2017simple}, scribbles-level~\cite{wang2018interactive}, and even point-level~\cite{qu2020weakly,zhang2021interactive} labeling, using rough annotations as supervision. 
But these methods typically yield sub-optimal performance~\cite{zheng2019biomedical} because accurate delineation data are not available for model training. 
The fourth type annotates only ``worthy" samples that help improve the final segmentation accuracy, e.g., active learning methods~\cite{yang2017suggestive,shi2019active,dai2020suggestive,wang2021annotation}. 
However, active learning methods require medical experts to provide annotations interactively, often leading to a ``human-machine disharmony" problem. 
To address the annotation bottleneck, some studies~\cite{zheng2019biomedical,zheng2020annotation} managed to avoid human-machine iterations by selecting representative samples to annotate in one shot, but still needed to annotate a considerable amount of samples.

To 
reduce the annotation burden, two main types of mask propagation DL methods have been developed. The first type is slice reconstruction, which propagates an annotated 2D slice through the entire 3D volume by matching correspondences between consecutive slices~\cite{yeung2021sli2vol,wu2022self}.
The second type is slice registration, which propagates an annotated 2D slice throughout the 3D volume by establishing spatial transformations between consecutive slices~\cite{bitarafan2022vol2flow,bitarafan20203d,li2022pln,osman2022semi}.
For example, in Sli2Vol~\cite{yeung2021sli2vol}, a self-supervised method was proposed to propagate a provided labeled slice for segmentation. Specifically, it first learns correspondences from adjacent slices by solving a slice reconstruction task in the training stage. Then, in the test stage, the learned correspondences are leveraged to propagate a labeled slice to the other 
slices for segmentation.
Despite their success, these methods still suffer from several drawbacks: (i) They are prone to error drift (i.e., error accumulation) due to the inter-slice propagation of reconstruction/registration errors. (ii) They do not handle discontinuity well (e.g., unseen objects emerging or seen objects ending), which can occur between consecutive slices in 3D images, as they focus on exploiting object continuity.
Observe that, intuitively, by incorporating segmentation/pseudo-labels (PLs) to provide certain supervision for slice reconstruction in the training stage, the correspondences can be better learned and more reliable, as object discontinuity can be compensated by PLs.

In this work, we propose a new self-supervised mask propagation framework, called \proposed, {for segmenting any anatomical structures in 3D medical images by labeling only a single slice per training and testing volume.} 
Specifically, in the training stage, we first generate PLs for the training volumes by propagating a provided labeled slice to the other slices in each volume using Sli2Vol~\cite{yeung2021sli2vol}. 
Then, we introduce a new Object Estimation Guided Correspondence Flow Network (OEG-CFN) to learn reliable correspondences for subsequent propagation.
In the test stage, the learned correspondences are utilized to propagate a labeled slice to the other slices of a test volume. 
Unlike Sli2Vol~\cite{yeung2021sli2vol}, we design OEG-CFN to learn correspondences between both consecutive slices and corresponding PLs in a self-supervised manner.
Intuitively, the included PLs guide the model to focus on the estimated objects during the correspondence learning process, effectively addressing the error drift and discontinuity issues.
Extensive experiments on nine public datasets (both CT and MRI) covering ten different structures of interest (SOIs) show the effectiveness of our new method and the improved generalizability across different SOIs, modalities, and modals.

Our main contributions are three-fold:
(i) We propose a new self-supervised mask propagation framework for {segmenting any anatomical structures in 3D medical images using only a single annotated slice in each training and testing volume.}
(ii) We develop a new Object Estimation Guided Correspondence Flow Network (OEG-CFN) to learn reliable correspondences between consecutive slices and corresponding PLs in a self-supervised manner, effectively addressing the error drift and discontinuity issues.
(iii) Our method achieves significant improvements on nine public datasets (both CT and MRI) over known methods, and demonstrates better generalizability across different SOIs, modalities, and modals.

\begin{figure*}[t]
  \centering
   \includegraphics[width=0.87\linewidth]{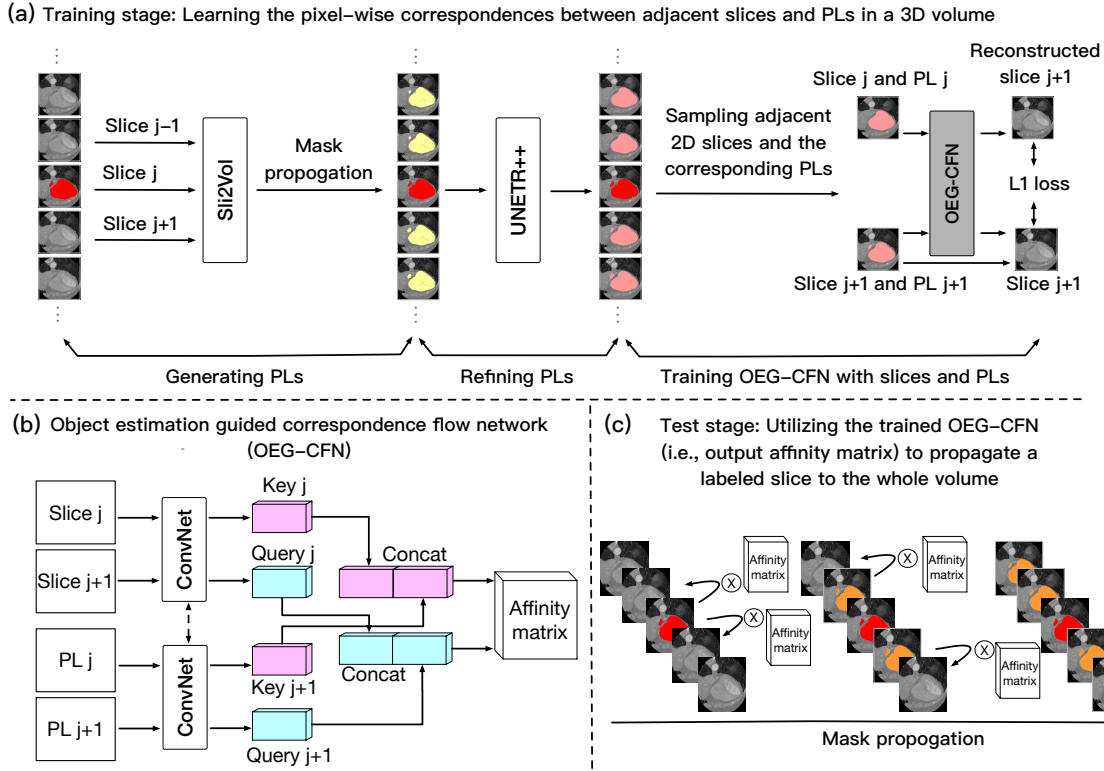}
   \caption{The pipeline of our proposed framework. (a) Training stage: Adjacent 2D slices and their corresponding generated pseudo-labels (PLs) are sampled from a 3D volume to train the Object Estimation Guided Correspondence Flow Network (OEG-CFN). (b) The architecture of OEG-CFN. (c) Test stage: The trained OEG-CFN is used to propagate a labeled slice to the other slices of the entire volume (five slices of a test volume are shown in the example). Red annotations represent ground truth segmentations, yellow and pink annotations represent PLs, and orange annotations represent the final segmentations. $\bigotimes$ denotes matrix multiplication.}
   \label{fig:pipeline}
\end{figure*}

\section{Related Work}
\label{sec:related}
To alleviate the burden of manual annotation and enhance generalizability in 3D
medical image segmentation, mask propagation DL methods have been proposed, including slice reconstruction-based methods~\cite{yeung2021sli2vol,wu2022self} and slice registration-based methods~\cite{bitarafan2022vol2flow,bitarafan20203d,li2022pln,osman2022semi}. 

Different annotation scenarios were considered using the known mask propagation DL methods. One scenario involves labeling one slice per training volume (e.g., \cite{bitarafan20203d}) without any labels on test volumes. However, these methods may not work well with new objects in test volumes.
Another scenario involves labeling one slice per test volume (e.g., \cite{yeung2021sli2vol,bitarafan2022vol2flow}) without using any labels of training volumes. In particular, Sli2Vol~\cite{yeung2021sli2vol} uses only a single annotated slice per test volume. But these methods do not leverage training labels and may not perform well.
We combine the above two scenarios, labeling one slice per training volume and one slice per test volume.
Compared to~\cite{yeung2021sli2vol,bitarafan2022vol2flow}, our \proposed requires an additional amount of very sparse annotations (i.e., a single annotated slice per training volume).

{Incorporating information between nearby slices in volumetric data can be achieved in various ways. For example, CSA-Net~\cite{csa-net} used pixel-level cross-slice attention to enhance the segmentation of a central slice, while CSAM~\cite{csam} employed slice-level attention across feature maps at multiple scales. Both these methods followed a standard 2.5D approach by stacking neighboring slices. However, compared to existing mask propagation approaches, these methods tend to be less memory efficient.}

\section{Method}
\label{sec:method}

\subsection{Problem Formulation and Method Overview}
\label{subsec:ovewview}
\textbf{Problem Formulation.} Let $\mathbb{X}_{train} = \{X_1, X_2, \ldots$, $X_N\}$ be a given set of $N$ 3D training images, where each volume $X_i$ contains $D$ 2D slices, $X_i = (S_{1}^{i}, S_{2}^{i}, \ldots, S_{D}^{i})$, and only a single annotated slice is provided for each $X_i$.
Given a 3D testing image $X_{test} = (S'_{1}, S'_{2}, \ldots, S'_{D'})$, 
our goal is to segment the SOIs in the test volume with a given 2D segmentation mask of a single slice $S'_{j}$ in $X_{test}$.

\textbf{Overview of the Method.}
As shown in Fig.~\ref{fig:pipeline}, our \proposed pipeline consists of the following main steps. In the training stage, we first generate and refine PLs for all the training volumes. Then, we learn the pixel-wise correspondences between adjacent slices and the corresponding PLs in a self-supervised manner using the proposed Object Estimation Guided Correspondence Flow Network (OEG-CFN).  In the test stage, the trained OEG-CFN is used to propagate an annotated slice to the other slices in the test volume, generating segmentation of the entire volume.

\subsection{Correspondence Learning with Object Estimation Guided Correspondence Flow Network}
\label{subsec:flow-network}
{\textbf{Review of Sli2Vol.} 
Sli2Vol~\cite{yeung2021sli2vol} is an interesting and closely related mask propagation DL method, which proposed a
self-supervised approach for learning dense correspondences by matching neighboring slices via slice reconstruction. The learned correspondences (i.e., a set of affinity matrices) are then used for mask propagation by weighting and copying pixels between consecutive slices in the test stage.
Specifically, in the training stage, pairs of adjacent slices sampled from a training volume, $\{{\mathbf{S}_j}, {\mathbf{S}_{j+1}}\}$, are fed to a CNN network, ConvNet (parameterized by $\psi(\cdot;\theta)$), to learn  features of the key $\mathbf{k}_j$ and query $\mathbf{q}_{j+1}$, as:
\begin{equation}
[\mathbf{k}_j,\ \mathbf{q}_{j+1}] = [\psi(g(\mathbf{S}_j) ;\ \theta),\ \psi(g(\mathbf{S}_{j+1}) ;\ \theta)],
\end{equation}
where $g(\cdot)$ denotes an edge profile generator for the information bottleneck.
Then, an affinity matrix $\mathbf{A}_{j \rightarrow j+1}$ is computed from $\mathbf{k}_j$ and $\mathbf{q}_{j+1}$ to represent the feature similarity between slices ${\mathbf{S}_j}$ and ${\mathbf{S}_{j+1}}$, as:
\begin{equation}\label{eq-2}
\mathbf{A}_{j \rightarrow j+1}(u, v) = \frac{\exp\langle \mathbf{q}_{j+1}(u,:), \mathbf{k}_j(v,:) \rangle}{\sum_{p \in \Omega}{\exp\langle \mathbf{q}_{j+1}(u,:), \mathbf{k}_j(p,:) \rangle}},
\end{equation}
where $\langle\cdot{,}\cdot\rangle$ denotes the dot product of two vectors, and $\Omega$ is a window surrounding pixels of $v$ for computing local attention. In the test stage, the affinity matrices thus computed are leveraged to propagate the given mask of a single slice to the other slices for the segmentation of a test volume.}

Sli2Vol~\cite{yeung2021sli2vol} focuses on exploiting object continuity, i.e., it is capable of propagating the labels of a slice to its neighboring slices well when there are no drastic changes between these slices.
But, this assumption is often not held in 3D medical images, where discontinuities may occur between consecutive slices: 
Unseen objects 
may emerge, or seen objects may end.
This gives rise to error drift and discontinuity issues, and may incur sub-optimal performance.

Our idea is that by including segmentation/PLs to provide certain supervision for the slice reconstruction, the correspondences can be better learned to handle these issues. Hence, 
we propose a new self-supervised approach that consists of the following steps: PLs generation, PLs refinement, and correspondence learning with an Object Estimation Guided Correspondence Flow Network (OEG-CFN). 

\textbf{PLs Generation.} Given a training volume $X = (S_{1}, S_{2}$, $\ldots, S_{D})$ and the annotation $Y_{j}$ of a single slice $S_{j}$, we apply Sli2Vol~\cite{yeung2021sli2vol} to propagate the annotation $Y_{j}$ to the other slices, generating PLs for the whole volume $X$.
However, Sli2Vol~\cite{yeung2021sli2vol} neglects global 3D information within the whole volume as it determines the correspondences using only adjacent slices without considering the entire context.

\textbf{PLs Refinement.} To address this issue, we utilize a SOTA 3D model (i.e., UNETR++~\cite{shaker2022unetr++}) to refine the PLs. 
Specifically, we use the generated PLs of all the training slices to train UNETR++\cite{shaker2022unetr++}. Then, we apply the trained UNETR++ to the entire training volume $X$ to refine the PLs. By considering the 3D information of $X$, the quality of the PLs is improved. 
We experimentally choose UNETR++\cite{shaker2022unetr++} as our 3D model because it yields the best performance for medical image segmentation (see Table~\ref{tab:unetr++}).


\begin{table}
\centering
{\small{
{
\begin{tabular}{lcc}\hline
\multirow{2}{*}{Method}  & \multicolumn{2}{c}{Dice}  \\
& Decath-spleen & Decath-liver  \\\hline
 TransUNet~\cite{chen2021transunet} & \begin{tabular}[c]{@{}l@{}}0.950 $\pm$ 0.013\end{tabular}  & \begin{tabular}[c]{@{}l@{}}0.944 $\pm$ 0.020\end{tabular}  \\
    CoTr~\cite{xie2021cotr} & \begin{tabular}[c]{@{}l@{}}0.954 $\pm$ 0.018\end{tabular} & \begin{tabular}[c]{@{}l@{}}0.942 $\pm$ 0.014\end{tabular}\\
   UNETR~\cite{hatamizadeh2022unetr} & \begin{tabular}[c]{@{}l@{}}0.964 $\pm$ 0.016\end{tabular}&\begin{tabular}[c]{@{}l@{}}0.961 $\pm$ 0.015\end{tabular} \\
    UNETR++~\cite{shaker2022unetr++} &\begin{tabular}[c]{@{}l@{}}\textbf{0.971} $\pm$ 0.012\end{tabular}& \begin{tabular}[c]{@{}l@{}}\textbf{0.964} $\pm$ 0.010\end{tabular}\\   \hline     
\end{tabular}}
\caption{{Segmentation results of different models on the Decath-spleen and liver datasets~\cite{simpson2019large}. The best results are marked in \textbf{bold}.}}\label{tab:unetr++}
}
}
\end{table}

\textbf{Object Estimation Guided Correspondence Flow Network (OEG-CFN).}
To include PLs for providing supervision for slice reconstruction, we propose a new Object Estimation Guided Correspondence Flow Network (OEG-CFN) to learn reliable correspondences. Our idea is to decompose the key and query features into two sets so that those features learned from consecutive slices handle continuity, and the features learned from PLs handle discontinuity.
As shown in Fig.~\ref{fig:pipeline}(b), OEG-CFN has two paths. The top path uses a ConvNet to learn the key and query features from consecutive slices, while the bottom path learns another two sets of key and query features using PLs with a ConvNet. Both the ConvNets share the same architecture and parameters. The key and query features learned from the slices and PLs are then concatenated respectively to form the final key and query features, which are subsequently used to compute the affinity matrix as in Eq.~(\ref{eq-2}).

The PLs thus generated provide information on the estimated objects, helping our OEG-CFN deal with the discontinuity issue effectively and improve label propagation quality by learning correspondences from the PLs.

\subsection{Gradient Enhanced Image Generator}
\label{subsec:GEI}
The model seeks to learn reliable correspondences by solving the slice reconstruction pre-text task~\cite{yeung2021sli2vol}. In order to achieve this, the slice reconstruction pre-text task should not be solved merely in a simple way (e.g., by simply matching the pixel intensities of two neighboring slices). 
Sli2Vol~\cite{yeung2021sli2vol} proposed an edge profile generator (i.e., computing the first-order derivatives of each pixel's intensity value) as an information bottleneck to avoid trivial solutions. This encourages the model to focus more on the edges during slice reconstruction.
However, it still has several drawbacks: (1) The edge profile generator that utilizes first-order derivatives is highly sensitive to noise; small variations in pixel intensity can cause significant changes in the derivative values, leading to sub-optimal performance. (2) It can detect regions accurately only with high-intensity changes, resulting in inaccurate representation of the edges.

To address these issues, we propose to generate {\it gradient-enhanced images}. Specifically, given a slice, we first transform the intensity value of each pixel into a normalized histogram of second-order derivatives, computed in $d$ different directions and at $s$ different scales, and then apply softmax normalization across all the derivative values, as:
\begin{equation}
G(s,d,p) = \text{softmax} ( \frac{\partial^2}{\partial x_{1,1}^2} I(p), \frac{\partial^2}{\partial x_{1,2}^2} I(p), \ldots, \frac{\partial^2}{\partial x_{(d,s)}^2} I(p) ), \
\end{equation}
where \( I(p) \) denotes the intensity value of a pixel \( p \), and \( \partial^2 / \partial x_{i,j}^2 \) is the second-order derivative along direction \( i \) at scale \( j \). 
{Here, scales refer to the window sizes used to compute the second-order derivative of the center pixel in the window.}
After that, we concatenate these values with the intensity values to form a gradient-enhanced image. We refer to this as the gradient-enhanced image generator (GEIG).

Using the gradient-enhanced images, the model learns the correspondences. This method is designed to enable the model to learn more reliable correspondences during slice reconstruction because of several benefits of the gradient-enhanced images: (1) They are capable of mitigating noise sensitivity by considering the rate of change of the gradient rather than the intensity values; (2) they enable more precise edge localization by detecting zero-crossings, which correspond to the actual positions of the edges.

\subsection{Inference with Mask Propagation}
\label{subsec:infer}
Our \proposed takes a pair of slices as input and finds the correspondences that map one slice to the other. It is trained on all pairs of consecutive training slices. 
Once trained, in the test stage, given two consecutive slices $x,x'$, it can compute the correspondences that map $x$ to its neighboring slice $x'$.
Thus, the mask of $x'$ can be estimated by applying the correspondences to the mask of $x$, allowing the masks of the neighboring slices of a labeled slice to be generated.

Specifically, given a test volume $X_{test} = (S'_{1}, S'_{2}$, $\ldots, S'_{D'})$, two consecutive slices ${{S'}_i}$ and ${{S'}_{i+1}}$ are sampled from $X_{test}$.
The gradient-enhanced images are then computed using GEIG (Section~\ref{subsec:GEI}), and these images are fed to the trained model to obtain the affinity matrix $\mathbf{A}_{i \rightarrow i+1}$. 
This matrix is then used to propagate a mask $\mathbf{\hat{M}}_i$ of slice $S'_i$ to generate the mask $\mathbf{\hat{M}}_{i+1}$ of slice $S'_{i+1}$, as:
\begin{equation}
\hat{\mathbf{M}}_{i+1}(u,:) = \sum_v{\mathbf{A}_{i \rightarrow i+1}(u,v)\mathbf{\hat{M}}_i(v,:)}.
\end{equation}

Note that, unlike in the training stage, no PLs of the slices are needed for computing the affinity matrix during the test stage.
This process of computations is then repeated for another two consecutive slices, say
${\mathbf{S'}_{i+1}}$ and ${\mathbf{S'}_{i+2}}$, 
in either direction until the whole test volume is covered (an example is shown in Fig.~\ref{fig:pipeline}(c)).

\begin{table*}[t]
\scalebox{0.68}{%
\begin{tabular}{|l|l|l|l|l|l|l|l|l|l|l|l|l|l|l|l|c|}
\hline
\multicolumn{1}{|c|}{\textit{{Modality}}} & 
\multicolumn{14}{c|}{\textit{{Abdominal and Chest CT}}} &                                  
\\ \cline{1-15}
\multicolumn{1}{|c|}{\textit{{\begin{tabular}[c]{@{}c@{}}Training Dataset \\ (for rows (e) to (i))\end{tabular}}}}                                        & 
\multicolumn{14}{c|}{\textit{{C4KC-KiTS, CT-LN, and CT-Pancreas}}}             &   
\\ \cline{1-15}
\multicolumn{1}{|c|}{\textit{{\begin{tabular}[c]{@{}c@{}}Testing Dataset\end{tabular}}}}                              & 
\multicolumn{1}{|c|}{\textit{{\begin{tabular}[c]{@{}l@{}}Decath-\\ Hea (MRI)\end{tabular}}}} &
\textit{{Sliver07}}                                       & \textit{{CHAOS}}                                          & \textit{{\begin{tabular}[c]{@{}l@{}}Decath-\\ Liv\end{tabular}}} & \textit{{\begin{tabular}[c]{@{}l@{}}Decath-\\ Spl\end{tabular}}} & \textit{{\begin{tabular}[c]{@{}l@{}}Decath-\\ Pan\end{tabular}}} & \multicolumn{8}{c|}{\textit{{3D-IRCADb-01 and 3D-IRCADb-02}}}    &     

\\ \cline{1-15}
\multicolumn{1}{|c|}{\textit{{ROI}}}                                          & 
\textit{{LA}}                                          &
\textit{{Liv}}                                          & \textit{{Liv}}                                          & \textit{{Liv}}                                                   & \textit{{Spl}}                                                   & \textit{{Pan}}                                                   & \textit{{Hea}}                                           & \textit{{\begin{tabular}[c]{@{}l@{}}Gal\end{tabular}}} & 
\textit{{Kid}}                                         & \textit{{\begin{tabular}[c]{@{}l@{}}Sur\end{tabular}}} & \textit{{Liv}}                                          & \textit{{Lun}}                                            & \textit{{Pan}}                                       & 
\textit{{Spl}}                                         & \textit{{\begin{tabular}[c]{@{}c@{}}Mean\end{tabular}}} 

\\ \cline{1-15}
\multicolumn{1}{|c|}{\textit{{\# of Volumes}}}                           & 
\textit{{20}} &
 \textit{{20}}                                             & \textit{{20}}                                             & \textit{{131}}                                                     & \textit{{41}}                                                       & \textit{{281}}                                                        & \textit{{3}}                                               & \textit{{8}}                                                       & \textit{{17}}                                             & \textit{{11}}                                                        & \textit{{22}}                                             & \textit{{12}}                                              & \textit{{4}}                                              & 
\textit{{7}}                                              &   

\\ \hline
\multicolumn{16}{|c|}{\cellcolor[HTML]{EFEFEF} Trained with Fully Annotated Data} 
\\ \hline
\begin{tabular}[c]{@{}l@{}}(a) FS-SD\end{tabular}          &
\begin{tabular}[c]{@{}l@{}}92.7\cite{isensee2021nnu} \\(\textbf{94.4})\end{tabular}            & 
\begin{tabular}[c]{@{}l@{}}94.8\cite{ahmad2019deep} \\(\textbf{96.2})\end{tabular}           & 
\begin{tabular}[c]{@{}l@{}}\textbf{97.8}\cite{kavur2020chaos} \\(96.4)\end{tabular}                & 
\begin{tabular}[c]{@{}l@{}}95.4\cite{isensee2021nnu}\\ (\textbf{95.9})\end{tabular}                                                               & 
\begin{tabular}[c]{@{}l@{}}96.0\cite{isensee2021nnu} \\ (\textbf{97.1})\end{tabular}                                                               & 
\begin{tabular}[c]{@{}l@{}}79.3\cite{isensee2021nnu} \\ (\textbf{79.5})\end{tabular}                                             & 
\begin{tabular}[c]{@{}l@{}}\multicolumn{1}{c}{-} \\ (\textbf{97.7})\end{tabular}                                                 & 
\begin{tabular}[c]{@{}l@{}}\multicolumn{1}{c}{-} \\ (\textbf{72.4})\end{tabular}                                           & 
\begin{tabular}[c]{@{}l@{}}\multicolumn{1}{c}{-} \\ (\textbf{97.1}  )\end{tabular}                                                     & 
\begin{tabular}[c]{@{}l@{}}\multicolumn{1}{c}{-} \\ (\textbf{69.7})\end{tabular}  & 
\begin{tabular}[c]{@{}l@{}}\textbf{96.5}\cite{tran2020multiple} \\ (\textbf{96.5})\end{tabular}                               & 
\begin{tabular}[c]{@{}l@{}}\multicolumn{1}{c}{-} \\ (\textbf{96.9})\end{tabular}                                            & 
\begin{tabular}[c]{@{}l@{}}\multicolumn{1}{c}{-} \\ (\textbf{72.4})\end{tabular}                                           &
\begin{tabular}[c]{@{}l@{}}\multicolumn{1}{c}{-} \\ (\textbf{94.2})\end{tabular}   &
\begin{tabular}[c]{@{}l@{}}\multicolumn{1}{c}{-} \\ (\textbf{89.7})\end{tabular} 
\\ \hline
\begin{tabular}[c]{@{}l@{}}(b) FS-DD\end{tabular}     & 
\multicolumn{1}{c|}{-}                                            &
\begin{tabular}[c]{@{}l@{}}74.8\\ $\pm$13.2\end{tabular}         &
\begin{tabular}[c]{@{}l@{}}76.5\\ $\pm$8.8\end{tabular}          & 
\begin{tabular}[c]{@{}l@{}}56.0\\ $\pm$23.6\end{tabular}                  & \multicolumn{1}{c|}{-}                                                     & \multicolumn{1}{c|}{-}                                                       & \multicolumn{1}{c|}{-}                                            & 
\multicolumn{1}{c|}{-}                                                    & \multicolumn{1}{c|}{-}                                           & 
\multicolumn{1}{c|}{-}                                                      & \multicolumn{1}{c|}{-}                                           &
\multicolumn{1}{c|}{-}                                            & 
\multicolumn{1}{c|}{-}                                           & 
\multicolumn{1}{c|}{-}                                           & 
-                                                                        
\\ \hline
\multicolumn{16}{|c|}{\cellcolor[HTML]{EFEFEF} Semi-supervised}                                       
\\ \hline
\begin{tabular}[c]{@{}l@{}}(c) FS-SS\end{tabular}         & 
\begin{tabular}[c]{@{}l@{}}47.7\\ $\pm$6.8\end{tabular}          &
\begin{tabular}[c]{@{}l@{}}85.3\\ $\pm$5.1\end{tabular}          & 
\begin{tabular}[c]{@{}l@{}}79.1\\ $\pm$6.4\end{tabular}          & 
\begin{tabular}[c]{@{}l@{}}84.6\\ $\pm$2.7\end{tabular}                   & \begin{tabular}[c]{@{}l@{}}75.6\\ $\pm$9.6\end{tabular}                   & \begin{tabular}[c]{@{}l@{}}50.3\\ $\pm$11.5\end{tabular}                     & \begin{tabular}[c]{@{}l@{}}23.3\\ $\pm$6.9\end{tabular}           & 
\begin{tabular}[c]{@{}l@{}}44.6\\ $\pm$16.1\end{tabular}                  & \begin{tabular}[c]{@{}l@{}}58.4\\ $\pm$14.3\end{tabular}         & 
\begin{tabular}[c]{@{}l@{}}27.7\\ $\pm$16.3\end{tabular}                    & \begin{tabular}[c]{@{}l@{}}81.2\\ $\pm$8.2\end{tabular}         & 
\begin{tabular}[c]{@{}l@{}}80.8\\ $\pm$9.9\end{tabular}          & 
\begin{tabular}[c]{@{}l@{}}18.2\\ $\pm$6.7\end{tabular}          & 
\begin{tabular}[c]{@{}l@{}}59.2\\ $\pm$5.4\end{tabular}          & 
58.3                                                                   
\\ \hline
\begin{tabular}[c]{@{}l@{}}(d) VM~\cite{balakrishnan2019voxelmorph}\end{tabular}                     & 
\begin{tabular}[c]{@{}l@{}}39.5\\ $\pm$7.6\end{tabular}          & 
\begin{tabular}[c]{@{}l@{}}57.2\\ $\pm$9.8\end{tabular}          & 
\begin{tabular}[c]{@{}l@{}}66.5\\ $\pm$10.5\end{tabular}         & 
\begin{tabular}[c]{@{}l@{}}38.5\\ $\pm$12.5\end{tabular}                  & \begin{tabular}[c]{@{}l@{}}61.5\\ $\pm$19.5\end{tabular}                   & \begin{tabular}[c]{@{}l@{}}21.4\\ $\pm$6.7\end{tabular}                      & \begin{tabular}[c]{@{}l@{}}20.3\\ $\pm$6.5\end{tabular}           & 
\begin{tabular}[c]{@{}l@{}}20.2\\ $\pm$12.2\end{tabular}                  & \begin{tabular}[c]{@{}l@{}}70.1\\ $\pm$18.6\end{tabular}         & 
\begin{tabular}[c]{@{}l@{}}41.1\\ $\pm$15.3\end{tabular}                    & \begin{tabular}[c]{@{}l@{}}60.5\\ $\pm$9.7\end{tabular}          & 
\begin{tabular}[c]{@{}l@{}}38.7\\ $\pm$21.2\end{tabular}          & 
\begin{tabular}[c]{@{}l@{}}28.3\\ $\pm$11.0\end{tabular}         & 
\begin{tabular}[c]{@{}l@{}}54.1\\ $\pm$12.4\end{tabular}         & 
\begin{tabular}[c]{@{}l@{}}41.1\end{tabular}                                                                      
\\ \hline
        
\begin{tabular}[c]{@{}l@{}}(e) Sli2Vol~\cite{yeung2021sli2vol} \end{tabular} & 
\begin{tabular}[c]{@{}l@{}}51.6\\ $\pm$8.2\end{tabular} & 
\begin{tabular}[c]{@{}l@{}}87.9\\ $\pm$6.3\end{tabular} & \begin{tabular}[c]{@{}l@{}}89.4\\ $\pm$3.1\end{tabular} & \begin{tabular}[c]{@{}l@{}}84.0\\ $\pm$8.8\end{tabular}          & \begin{tabular}[c]{@{}l@{}}88.7\\ $\pm$7.7\end{tabular} &                                            \begin{tabular}[c]{@{}l@{}}51.4\\ $\pm$10.6\end{tabular}      &                                   \begin{tabular}[c]{@{}l@{}}79.4\\ $\pm$7.3\end{tabular} & \begin{tabular}[c]{@{}l@{}}43.2\\ $\pm$24.3\end{tabular}    &      \begin{tabular}[c]{@{}l@{}}92.2\\ $\pm$5.8\end{tabular} & 
\begin{tabular}[c]{@{}l@{}}45.0\\ $\pm$15.6\end{tabular}                    & \begin{tabular}[c]{@{}l@{}}87.0\\ $\pm$3.6\end{tabular} & \begin{tabular}[c]{@{}l@{}}80.8\\ $\pm$6.2\end{tabular} & \begin{tabular}[c]{@{}l@{}}54.4\\ $\pm$6.4\end{tabular} & \begin{tabular}[c]{@{}l@{}}91.6\\ $\pm$4.0\end{tabular} & 
73.3
\\ \hline


\begin{tabular}[c]{@{}l@{}}(f) Vol2Flow~\cite{bitarafan2022vol2flow} \end{tabular} & 
\begin{tabular}[c]{@{}l@{}}51.1\\ $\pm$9.6\end{tabular} & 
\begin{tabular}[c]{@{}l@{}}\underline{92.1}\\ $\pm$4.8\end{tabular} & \begin{tabular}[c]{@{}l@{}}84.4\\ $\pm$4.1\end{tabular} & \begin{tabular}[c]{@{}l@{}}85.4\\ $\pm$6.5\end{tabular}          & \begin{tabular}[c]{@{}l@{}}88.7\\ $\pm$10.2\end{tabular}          & \begin{tabular}[c]{@{}l@{}}\underline{57.3}\\ $\pm$7.3\end{tabular}            & \begin{tabular}[c]{@{}l@{}}80.3\\ $\pm$6.9\end{tabular} & \begin{tabular}[c]{@{}l@{}}57.2\\ $\pm$16.4\end{tabular}          & \begin{tabular}[c]{@{}l@{}}88.4\\ $\pm$6.2\end{tabular} & 
\begin{tabular}[c]{@{}l@{}}42.3\\ $\pm$14.4\end{tabular}                    & \begin{tabular}[c]{@{}l@{}}\underline{88.6}\\ $\pm$2.7\end{tabular} & \begin{tabular}[c]{@{}l@{}}83.5\\ $\pm$2.2\end{tabular} & \begin{tabular}[c]{@{}l@{}}\underline{62.4}\\ $\pm$9.5\end{tabular} & \begin{tabular}[c]{@{}l@{}}87.4\\ $\pm$7.5\end{tabular} & 
74.9
\\ \hline


\proposed                                                                          & \multicolumn{15}{|c|}{\textit{\textbf{Ablation Study}}}                                          
\\ \hdashline
\begin{tabular}[c]{@{}l@{}}(g) PLs \\+ OEG-CFN\end{tabular}                            & 
\begin{tabular}[c]{@{}l@{}}52.3\\ $\pm$8.4\end{tabular}          & 
\begin{tabular}[c]{@{}l@{}}88.2\\ $\pm$3.9\end{tabular}          & 
\begin{tabular}[c]{@{}l@{}}89.8\\ $\pm$2.3\end{tabular}          & 
\begin{tabular}[c]{@{}l@{}}85.3\\ $\pm$6.0\end{tabular}                  & \begin{tabular}[c]{@{}l@{}}90.2\\ $\pm$4.7\end{tabular}                   & \begin{tabular}[c]{@{}l@{}}50.6\\ $\pm$12.2\end{tabular}                     & \begin{tabular}[c]{@{}l@{}}88.1\\ $\pm$5.4\end{tabular}          & 
\begin{tabular}[c]{@{}l@{}}64.2\\ $\pm$8.4\end{tabular}                  & \begin{tabular}[c]{@{}l@{}}92.7\\ $\pm$4.6\end{tabular}         & 
\begin{tabular}[c]{@{}l@{}}49.6\\ $\pm$17.4\end{tabular}                    & \begin{tabular}[c]{@{}l@{}}86.2\\ $\pm$3.2\end{tabular}          & 
\begin{tabular}[c]{@{}l@{}}93.2\\ $\pm$5.4\end{tabular}          & 
\begin{tabular}[c]{@{}l@{}}52.4\\ $\pm$7.3\end{tabular}         & 
\begin{tabular}[c]{@{}l@{}}91.1\\ $\pm$3.6\end{tabular}         & 
76.7    

\\ \hdashline
\begin{tabular}[c]{@{}l@{}}(h) Refined PLs \\+ OEG-CFN\end{tabular}                & 
\begin{tabular}[c]{@{}l@{}}52.4\\ $\pm$7.3\end{tabular}          &
\begin{tabular}[c]{@{}l@{}}88.4\\ $\pm$3.3\end{tabular}          & 
\begin{tabular}[c]{@{}l@{}}90.6\\ $\pm$2.5\end{tabular}          & 
\begin{tabular}[c]{@{}l@{}}86.7\\ $\pm$7.9\end{tabular}                  & \begin{tabular}[c]{@{}l@{}}90.7\\ $\pm$3.5\end{tabular}                   & \begin{tabular}[c]{@{}l@{}}50.7\\ $\pm$10.3\end{tabular}                     & \begin{tabular}[c]{@{}l@{}}88.7\\ $\pm$4.9\end{tabular}          & 
\begin{tabular}[c]{@{}l@{}}65.1\\ $\pm$8.2\end{tabular}                  & \begin{tabular}[c]{@{}l@{}}93.2\\ $\pm$4.4\end{tabular}         & \begin{tabular}[c]{@{}l@{}}50.1\\ $\pm$17.1\end{tabular}           & \begin{tabular}[c]{@{}l@{}}86.8\\ $\pm$3.3\end{tabular}          & 
\begin{tabular}[c]{@{}l@{}}94.5\\ $\pm$2.0\end{tabular}          & 
\begin{tabular}[c]{@{}l@{}}52.2\\ $\pm$7.9\end{tabular}          & 
\begin{tabular}[c]{@{}l@{}}91.8\\ $\pm$2.7\end{tabular}         & 
77.3

\\ \hdashline
\begin{tabular}[c]{@{}l@{}}(i) Refined PLs \\+ OEG-CFN \\+ GEIG (Ours)\end{tabular}              & 
{\begin{tabular}[c]{@{}l@{}}\underline{54.9}\\ $\pm$7.4\end{tabular}}          & 
\begin{tabular}[c]{@{}l@{}}88.7\\ $\pm$5.1\end{tabular}          & 
{\begin{tabular}[c]{@{}l@{}}\underline{91.2}\\ $\pm$3.0\end{tabular}}          & 
{\begin{tabular}[c]{@{}l@{}}\underline{88.4}\\ $\pm$5.6\end{tabular}}                   & {\begin{tabular}[c]{@{}l@{}}\underline{92.6}\\ $\pm$3.3\end{tabular}}                   & \begin{tabular}[c]{@{}l@{}}52.0\\ $\pm$10.1\end{tabular}                     & {\begin{tabular}[c]{@{}l@{}}\underline{90.6}\\ $\pm$2.9\end{tabular}}          & 
{\begin{tabular}[c]{@{}l@{}}\underline{68.7}\\ $\pm$7.0\end{tabular}}                  & {\begin{tabular}[c]{@{}l@{}}\underline{94.2}\\ $\pm$2.1\end{tabular}}         & 
{\begin{tabular}[c]{@{}l@{}}\underline{52.0}\\ $\pm$17.2\end{tabular}}                    & \begin{tabular}[c]{@{}l@{}}87.4\\ $\pm$2.8\end{tabular}          & 
{\begin{tabular}[c]{@{}l@{}}\underline{96.3}\\ $\pm$1.9\end{tabular}}          & 
\begin{tabular}[c]{@{}l@{}}54.4\\ $\pm$6.1\end{tabular}         & 
{\begin{tabular}[c]{@{}l@{}}\underline{92.8}\\ $\pm$2.8\end{tabular}}         & 
\underline{78.9}                                                                     
\\ \hline
\end{tabular}
}
\caption{Results (mean Dice scores $\pm$ standard deviation) of different methods on various datasets. \textbf{Row (a)} gives results of SOTA methods~\cite{isensee2021nnu, ahmad2019deep, kavur2020chaos, tran2020multiple} and UNETR++~\cite{shaker2022unetr++} trained by us (values in the brackets). The results in \textbf{row (a)} and \textbf{row (b)} are only partially available in the literature and are reported to demonstrate the approximate upper-bound and limitations of the fully supervised methods, which are not meant to be compared directly to our proposed approach. The best fully supervised results are marked in {\bf bold}. The best results of the semi-supervised methods are \underline{underlined}. The same for Table~\ref{table:result-mri}. 
}
\label{table:result-ct}
\end{table*}

\begin{table}[t]
\scalebox{0.7}{%
\begin{tabular}{|l|l|l|l|l|c|}
\hline
\multicolumn{1}{|c|}{\textit{{Modality}}} & 
\multicolumn{3}{c|}{\textit{{MRI}}} & 
                                          
\\ \cline{1-4}
\multicolumn{1}{|c|}{\textit{{\begin{tabular}[c]{@{}c@{}}Training Dataset \\ (for rows (e) to (i))\end{tabular}}}}                                        & 
\multicolumn{3}{|c|}{\textit{{\shortstack[l]{Brain Tumor (T1w modal)}}}} &   
\\ \cline{1-4}
\multicolumn{1}{|c|}{\textit{{\begin{tabular}[c]{@{}c@{}}Testing Dataset\end{tabular}}}}                              & 
\multicolumn{1}{|c|}{\textit{{\begin{tabular}[c]{@{}c@{}}FLAIR modal\end{tabular}}}} &
\multicolumn{1}{|c|}{\textit{{\begin{tabular}[c]{@{}c@{}}T1gd  modal\end{tabular}}}} &
\multicolumn{1}{|c|}{\textit{{\begin{tabular}[c]{@{}c@{}}T2w modal\end{tabular}}}} &   

\\ \cline{1-4}
\multicolumn{1}{|c|}{\textit{{ROI}}}                                          & 
\textit{{Tumor}}                                          &
\textit{{Tumor}}                                          &
\textit{{Tumor}}                                          
 & \textit{{\begin{tabular}[c]{@{}c@{}}Mean\end{tabular}}} 

\\ \cline{1-4}
\multicolumn{1}{|c|}{\textit{{\# of Volumes}}}                           &   
\textit{{266}} & 
\textit{{266}} & 
\textit{{266}} &

\\ \hline
\multicolumn{5}{|c|}{\cellcolor[HTML]{EFEFEF} Trained with Fully Annotated Data} 
\\ \hline
\begin{tabular}[c]{@{}l@{}}(a) FS-SD\end{tabular}          &
\begin{tabular}[c]{@{}l@{}}\textbf{93.6}\end{tabular}            &
\begin{tabular}[c]{@{}l@{}}\textbf{92.1}\end{tabular}            &
\begin{tabular}[c]{@{}l@{}}\textbf{93.3}\end{tabular}            &
\begin{tabular}[c]{@{}l@{}}\textbf{93.0}\end{tabular}            
\\ \hline

\begin{tabular}[c]{@{}l@{}}(b) FS-DD \end{tabular} & 
\begin{tabular}[c]{@{}l@{}}72.6 $\pm$11.7\end{tabular} & 
\begin{tabular}[c]{@{}l@{}}72.3 $\pm$12.5\end{tabular} & 
\begin{tabular}[c]{@{}l@{}}69.2 $\pm$14.8\end{tabular} & 
\begin{tabular}[c]{@{}l@{}}71.4 \end{tabular} 
\\ \hline


\multicolumn{5}{|c|}{\cellcolor[HTML]{EFEFEF}Semi-supervised}                                       
\\ \hline
\begin{tabular}[c]{@{}l@{}}(c) FS-SS\end{tabular}         & 
\begin{tabular}[c]{@{}l@{}}44.6 $\pm$7.9\end{tabular}          &
\begin{tabular}[c]{@{}l@{}}45.8 $\pm$7.2\end{tabular}          &
\begin{tabular}[c]{@{}l@{}}43.3 $\pm$8.6\end{tabular}          &
\begin{tabular}[c]{@{}l@{}}44.6\end{tabular} 
\\ \hline
\begin{tabular}[c]{@{}l@{}}(d) VM~\cite{balakrishnan2019voxelmorph}\end{tabular}                     & 
\begin{tabular}[c]{@{}l@{}}35.9 $\pm$8.4\end{tabular}          & 
\begin{tabular}[c]{@{}l@{}}36.5 $\pm$12.4\end{tabular}          & 
\begin{tabular}[c]{@{}l@{}}34.4 $\pm$10.7\end{tabular}          & 
    \begin{tabular}[c]{@{}l@{}}35.6  \end{tabular}
\\ \hline
        
\begin{tabular}[c]{@{}l@{}}(e) Sli2Vol~\cite{yeung2021sli2vol} \end{tabular} & 
\begin{tabular}[c]{@{}l@{}}49.5 $\pm$7.8\end{tabular} & 
\begin{tabular}[c]{@{}l@{}}50.9 $\pm$6.6\end{tabular} & 
\begin{tabular}[c]{@{}l@{}}51.3 $\pm$7.5\end{tabular} & 
\begin{tabular}[c]{@{}l@{}} 50.6 \end{tabular}
\\ \hline


\begin{tabular}[c]{@{}l@{}}(f) Vol2Flow~\cite{bitarafan2022vol2flow} \end{tabular} & 
\begin{tabular}[c]{@{}l@{}}48.8 $\pm$9.7\end{tabular} &
\begin{tabular}[c]{@{}l@{}}51.0 $\pm$10.1\end{tabular} &
\begin{tabular}[c]{@{}l@{}}52.1 $\pm$8.8\end{tabular} &
\begin{tabular}[c]{@{}l@{}} 50.6 \end{tabular}
\\ \hline


\proposed                                                                          & \multicolumn{5}{|c|}{\textit{\textbf{Ablation Study}}}                                          
\\ \hdashline
\begin{tabular}[c]{@{}l@{}}(g) PLs + OEG-CFN\end{tabular}                            & 
\begin{tabular}[c]{@{}l@{}}53.9 $\pm$7.4\end{tabular}          & 
\begin{tabular}[c]{@{}l@{}}51.6 $\pm$7.1\end{tabular}          & 
\begin{tabular}[c]{@{}l@{}}51.9 $\pm$6.2\end{tabular}          & 
\begin{tabular}[c]{@{}l@{}} 52.5 \end{tabular}

\\ \hdashline
\begin{tabular}[c]{@{}l@{}}(h) Refined PLs \\+ OEG-CFN\end{tabular}                & 
\begin{tabular}[c]{@{}l@{}}54.2 $\pm$7.7\end{tabular}          &
\begin{tabular}[c]{@{}l@{}}52.3 $\pm$7.2\end{tabular}          &
\begin{tabular}[c]{@{}l@{}}52.2 $\pm$6.7\end{tabular}          &
\begin{tabular}[c]{@{}l@{}} 52.9 \end{tabular}

\\ \hdashline
\begin{tabular}[c]{@{}l@{}}(i) Refined PLs \\+ OEG-CFN \\+ GEIG (Ours)\end{tabular}              & 
{\begin{tabular}[c]{@{}l@{}}\underline{54.9} $\pm$7.6\end{tabular}}          &
{\begin{tabular}[c]{@{}l@{}}\underline{53.6} $\pm$6.9\end{tabular}}          &
{\begin{tabular}[c]{@{}l@{}}\underline{53.7} $\pm$6.4\end{tabular}}          &
\begin{tabular}[c]{@{}l@{}} \underline{54.1} \end{tabular}
                                           
\\ \hline
\end{tabular}
}
\caption{Results of various methods on the MRI datasets with different modals. \textbf{Row (a)} gives results of UNETR++~\cite{shaker2022unetr++} trained by us.
The results in \textbf{row (a)} and \textbf{row (b)} are reported just to demonstrate the approximate upper-bound and limitations of the fully supervised methods,
which are not meant to be compared directly to our proposed approach. 
}
\label{table:result-mri}
\end{table}

\begin{table}
  \centering
  {\small{
  {
  \begin{tabular}{@{}lc@{}}
    \toprule
    Method & Dice \\
    \midrule
    SAM~\cite{kirillov2023segment} & 0.602 $\pm$ 0.024\\
    MedSAM~\cite{ma2023segment} & 0.774 $\pm$ 0.019\\
   ScribblePrompt~\cite{wong2024scribbleprompt} & \textbf{0.955} $\pm$ 0.007 \\
    \bottomrule
  \end{tabular}}
  }}
  \caption{{Segmentation results of different segmentation foundation models or interactive segmentation tools on
the Decath-Brain Tumours dataset~\cite{simpson2019large}.}}
  \label{tab:foundation}
\end{table}

\begin{table}[t]
\scalebox{0.7}{%
\begin{tabular}{|l|l|l|l|l|c|}
\hline
\multicolumn{1}{|c|}{\textit{{Modality}}} & 
\multicolumn{3}{c|}{\textit{{MRI}}} & 
                                          
\\ \cline{1-4}
\multicolumn{1}{|c|}{\textit{{\begin{tabular}[c]{@{}c@{}}Training Dataset \end{tabular}}}}                                        & 
\multicolumn{3}{|c|}{\textit{{\shortstack[l]{Brain Tumor (T1w modal)}}}} &   
\\ \cline{1-4}
\multicolumn{1}{|c|}{\textit{{\begin{tabular}[c]{@{}c@{}}Testing Dataset\end{tabular}}}}                              & 
\multicolumn{1}{|c|}{\textit{{\begin{tabular}[c]{@{}c@{}}FLAIR modal\end{tabular}}}} &
\multicolumn{1}{|c|}{\textit{{\begin{tabular}[c]{@{}c@{}}T1gd  modal\end{tabular}}}} &
\multicolumn{1}{|c|}{\textit{{\begin{tabular}[c]{@{}c@{}}T2w modal\end{tabular}}}} &   

\\ \cline{1-4}
\multicolumn{1}{|c|}{\textit{{ROI}}}                                          & 
\textit{{Tumor}}                                          &
\textit{{Tumor}}                                          &
\textit{{Tumor}}                                          
 & \textit{{\begin{tabular}[c]{@{}c@{}}Mean\end{tabular}}} 

\\ \cline{1-4}
\multicolumn{1}{|c|}{\textit{{\# of Volumes}}}                           &   
\textit{{266}} & 
\textit{{266}} & 
\textit{{266}} &


\\ \hline
\begin{tabular}[c]{@{}l@{}}(a) Sli2Vol+ (Ours)\end{tabular}              & 
{\begin{tabular}[c]{@{}l@{}}\textbf{54.9} $\pm$7.6\end{tabular}}          &
{\begin{tabular}[c]{@{}l@{}}\textbf{53.6} $\pm$6.9\end{tabular}}          &
{\begin{tabular}[c]{@{}l@{}}53.7 $\pm$6.4\end{tabular}}          &
\begin{tabular}[c]{@{}l@{}} \textbf{54.1} \end{tabular}
                                           
\\ \hline

\begin{tabular}[c]{@{}l@{}}(b) Sli2Vol+ (using \\ScribblePrompt~\cite{wong2024scribbleprompt})\end{tabular}              & 
{\begin{tabular}[c]{@{}l@{}}54.6 $\pm$7.2\end{tabular}}          &
{\begin{tabular}[c]{@{}l@{}}52.8 $\pm$7.3\end{tabular}}          &
{\begin{tabular}[c]{@{}l@{}}\textbf{53.9} $\pm$7.1\end{tabular}}          &
\begin{tabular}[c]{@{}l@{}} {53.8} \end{tabular}
                                           
\\ \hline

\end{tabular}
}
\caption{Results of various methods on the MRI datasets with different modals. In row (b), ScribblePrompt~\cite{wong2024scribbleprompt} is utilized to generate the single slice annotations of the training volumes.
}
\label{table:result-ScribblePromp}
\end{table}

\section{Experiments}
\label{sec:experi}

\subsection{Datasets}
\label{subsec:datasets}
We evaluate our proposed \proposed framework on nine public datasets (in CT and MRI) with ten different SOIs.

In the CT modality, we train our model on three datasets: C4KC-KiTS~\cite{c4kc}, CT-LN~\cite{lymphnodes}, and CT-Pancreas~\cite{pancreasct}, and test on seven different CT datasets: Sliver07~\cite{van07}, CHAOS~\cite{CHAOSdata2019}, 3Dircadb-01 and 3Dircadb-02~\cite{soler20103d}, and Decath-Spleen, Decath-Liver, and Decath-Pancreas~\cite{simpson2019large}, as well as one MRI dataset, Decath-Heart~\cite{simpson2019large}, spanning nine SOIs: Left atrium (LA), Liver (Liv), Spleen (Spl), Pancreas (Pan), Heart (Hea), Gallbladder (Gal), Kidney (Kid), Surrenal gland (Sur), and Lung (Lun).

In the MRI modality, we train our model on the Decath-Brain Tumours dataset~\cite{simpson2019large}, with the FLAIR, T1w, T1gd, and T2w modals. 
Specifically, we train our model on the T1w modal and test on the FLAIR, T1gd, and T2w modals.

We repeat the experiments five times with different random seeds and report the mean Dice scores $\pm$ standard deviation.

\subsection{Implementation Details}\label{imp}
Our experiments are conducted using the PyTorch library. Model training is performed on an NVIDIA A40 Graphics Card with 48GB GPU memory, utilizing the AdamW optimizer~\cite{loshchilov2017decoupled} with a weight decay of 0.005. The learning rate is set to 0.0001, and the number of training epochs is 4 for the experiments. The batch size for each case is set to the maximum size allowed by the GPU.

For UNETR++~\cite{shaker2022unetr++}, the learning rate is 0.01, with 1000 training epochs. The optimizer is SGD with a weight decay of 0.00003 and a momentum of 0.99.

\subsection{Baseline Comparison}
\label{subsec:baseline}
Following Sli2Vol~\cite{yeung2021sli2vol}, we compare our \proposed with three types of baselines.
(1) We compare two approaches trained on fully annotated 3D data, demonstrating the performance of SOTA fully supervised models with or without domain shifts. \textbf{Fully Supervised-Same Domain (FS-SD)} refers to the scenario where the training and testing data come from the same dataset. Results of SOTA methods~\cite{isensee2021nnu, ahmad2019deep, kavur2020chaos, tran2020multiple} and UNETR++~\cite{shaker2022unetr++} trained by us are reported. \textbf{Fully Supervised-Different Domain (FS-DD)} aims to evaluate the generalizability of fully supervised approaches when training and testing data come from different domains.
(2) We consider the scenario when only a single annotated slice is provided in each test volume to train a UNETR++\cite{shaker2022unetr++} model, referred to as \textbf{Fully Supervised-Single Slice (FS-SS)}. For example, in the CHAOS~\cite{CHAOSdata2019} dataset, the UNETR++ model is trained on 20 annotated slices (a single slice from each volume) and tested on the same set of 20 volumes. This approach utilizes the same amount of manual annotations as \proposed, to investigate whether a model trained on single slice annotations is sufficient to generalize to the whole volume.
(3) We compare with several SOTA mask propagation methods, including \textbf{VoxelMorph (VM)\cite{balakrishnan2019voxelmorph}, Sli2Vol\cite{yeung2021sli2vol}, and Vol2Flow~\cite{bitarafan2022vol2flow}}.

Following Sli2Vol~\cite{yeung2021sli2vol}, we randomly pick one of the $\pm 3$ slices around the slice with the largest ground truth (GT) annotation as an annotated slice.

\begin{figure*}[t]
  \centering
   \includegraphics[width=0.64\linewidth]{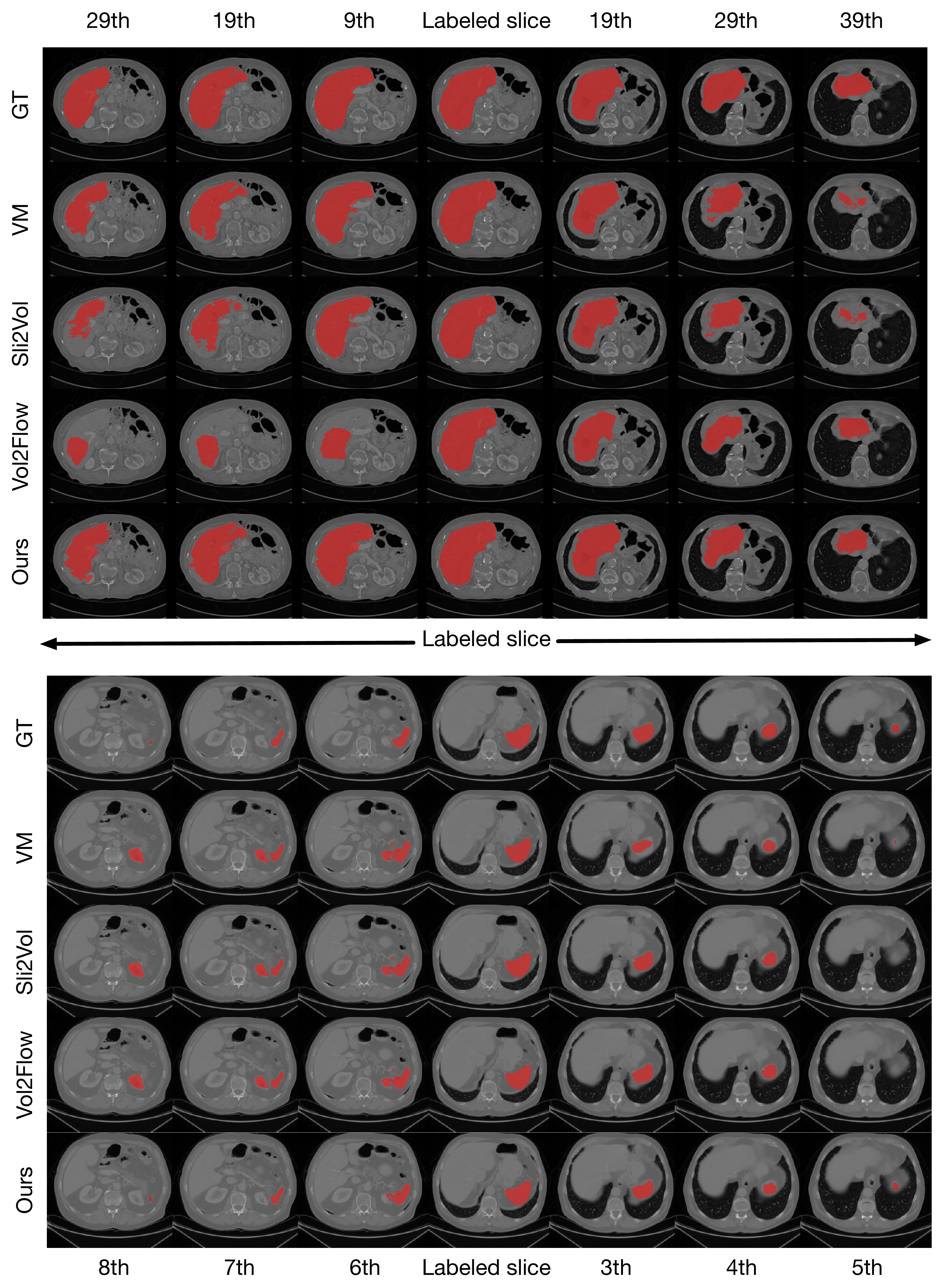}
   \caption{Examples of segmentation results by different methods on the Decath-Liver (top) and Decath-Spleen (bottom) datasets~\cite{simpson2019large}. In the examples, a labeled slice is propagated in two directions, and the ``$i$-th'' represents the position of a slice to which the labeled slice is propagated in that direction.}
   \label{fig:vis-1}
\end{figure*}

\section{Results and Discussions}
\label{subsec:result}

\subsection{Quantitative Results}
\label{subsec:quantitative}
{We compare our proposed \proposed, {which needs a single annotated slice in each training and testing volume, with two methods trained with fully annotated data 
and four semi-supervised methods 
which require only a single annotated slice per test volume.}

Table~\ref{table:result-ct} presents a quantitative comparison of various methods on datasets in both CT and MRI. From these results, we observe the following.
(1) Compared to the SOTA performance achieved by the Fully Supervised-Same Domain (FS-SD) (\textbf{row (a)}), a significant performance drop (over 20 in Dice score) can be seen for cross-domain evaluation (same SOIs, different datasets) (\textbf{row (b)}), by comparing \textbf{row (b)} with the values inside the brackets in \textbf{row (a)}. Yet, our method shows a drop of less than 7 Dice score, suggesting its potential for addressing domain shift problems.
(2) With the same amount of annotations (i.e., only a single annotated slice per test volume), our method (\textbf{row (i)}) significantly outperforms Fully Supervised-Single Slice (FS-SS) (\textbf{row (c)}) on all the datasets ($p<0.05$, t-test), with an average Dice score margin of over 20.
(3) For propagation-based methods (\textbf{rows (d)-(f)}), our method outperforms VM~\cite{balakrishnan2019voxelmorph}, Sli2Vol~\cite{yeung2021sli2vol}, and Vol2Flow~\cite{bitarafan2022vol2flow} on all the datasets. This suggests that our method incurs less severe error drifts thanks to its inclusion of PLs. Specifically, our method improves Sli2Vol~\cite{yeung2021sli2vol} by an average of 5.6 Dice score, demonstrating that our assumption (i.e., including PLs to provide supervision for slice reconstruction leads to better and more reliable correspondences) is valid.
(4) Our method outperforms FS-SS, VM~\cite{balakrishnan2019voxelmorph}, Sli2Vol~\cite{yeung2021sli2vol}, and Vol2Flow~\cite{bitarafan2022vol2flow} for cross-modality evaluation (i.e., different SOIs, different modalities), showing its better generalizability in cross-modality tasks.

Table~\ref{table:result-mri} gives a quantitative comparison of various methods on the MRI datasets with different modals. 
We use the T1w modal data for training and test the data with the FLAIR, T1gd, and T2w modals.
From these results, we observe the following.
(1) Our method significantly outperforms the FS-SS (\textbf{row (c)}) ($p<0.05$, t-test), with an average Dice score margin of over 9, using only a single annotated slice per volume. This demonstrates the effectiveness of our method in reducing the annotation burden.
(2) Our method outperforms the SOTA mask propagation methods (i.e., VM~\cite{balakrishnan2019voxelmorph}, Sli2Vol~\cite{yeung2021sli2vol}, and Vol2Flow~\cite{bitarafan2022vol2flow}) in cross-modal evaluation (i.e., same SOIs, different modals), showing its better generalizability across different modals.

\subsection{Qualitative Results}
\label{subsec:qualitative}
Fig.~\ref{fig:vis-1} presents some examples of segmentation results 
by VM~\cite{balakrishnan2019voxelmorph}, Sli2Vol~\cite{yeung2021sli2vol}, Vol2Flow~\cite{bitarafan2022vol2flow}, and our method on the Decath-Liver (top) and Decath-Spleen (bottom) datasets~\cite{simpson2019large}, with GT given as reference.

From the visual segmentation results in Fig.~\ref{fig:vis-1}, we can observe the following.
(1) When propagating to slices that are far away from the labeled slice, the segmentation results produced by our method are significantly better than those produced by the known mask propagation methods (e.g., see the segmentation results of the 29th and 39th slices). This validates that our method 
better handles the error accumulation issue during propagation.
(2) When propagating to slices where seen objects end, our method can still propagate well. For example, the spleen almost ends from the 7th slice to the 8th slice, yet our method generates accurate segmentation results. However, the other mask propagation methods generate false positives. This demonstrates that our method effectively deals with the discontinuity issue.

\subsection{Annotation Cost Analysis}
\label{subsec:comple}
Although our method requires an additional single slice annotation per training volume compared to Sli2Vol~\cite{yeung2021sli2vol}, which only needs a single slice annotation per test volume, the annotation efforts for our method are still relatively low (e.g., involving only a few hundred slices even for large training datasets, such as C4KC-KiTS~\cite{c4kc} with 300 training volumes). Considering the significant improvements it provides, the additional annotation efforts are worthwhile.

We would like to note that the additional annotation effort of \proposed can be further reduced by utilizing segmentation foundation models (e.g., SAM~\cite{kirillov2023segment} and MedSAM~\cite{ma2023segment}) or interactive biomedical image segmentation tools (e.g., ScribblePrompt~\cite{wong2024scribbleprompt}) to automatically generate annotations for training volumes.
We evaluate SAM~\cite{kirillov2023segment}, MedSAM~\cite{ma2023segment}, and ScribblePrompt~\cite{wong2024scribbleprompt} on the Decath-Brain Tumours dataset~\cite{simpson2019large}. Their quantitative segmentation results are reported in Table~\ref{tab:foundation}, and some visual results are shown in Fig.~\ref{fig:vis-2}. 
Since ScribblePrompt\cite{wong2024scribbleprompt} yielded the best performance in these experiments, we selected it to generate the annotations for the training volumes.

Table~\ref{table:result-ScribblePromp} reports the results of our method using segmentations generated by ScribblePrompt~\cite{wong2024scribbleprompt} (not the GT annotations). It achieves comparable performance.

\begin{figure}[t]
  \centering
   \includegraphics[width=0.83\linewidth]{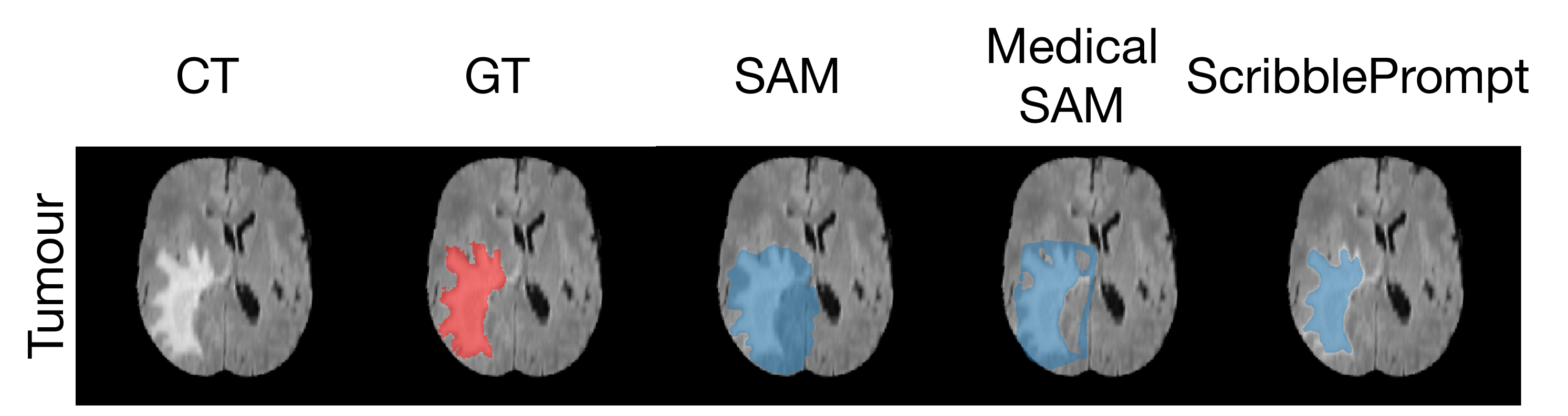}
   \caption{Examples of segmentation results by various segmentation foundation models or interactive segmentation tools on the Decath-Brain Tumours dataset~\cite{simpson2019large}.}
   \label{fig:vis-2}
\end{figure}

\begin{figure}[t]
  \centering
   \includegraphics[width=0.83\linewidth]{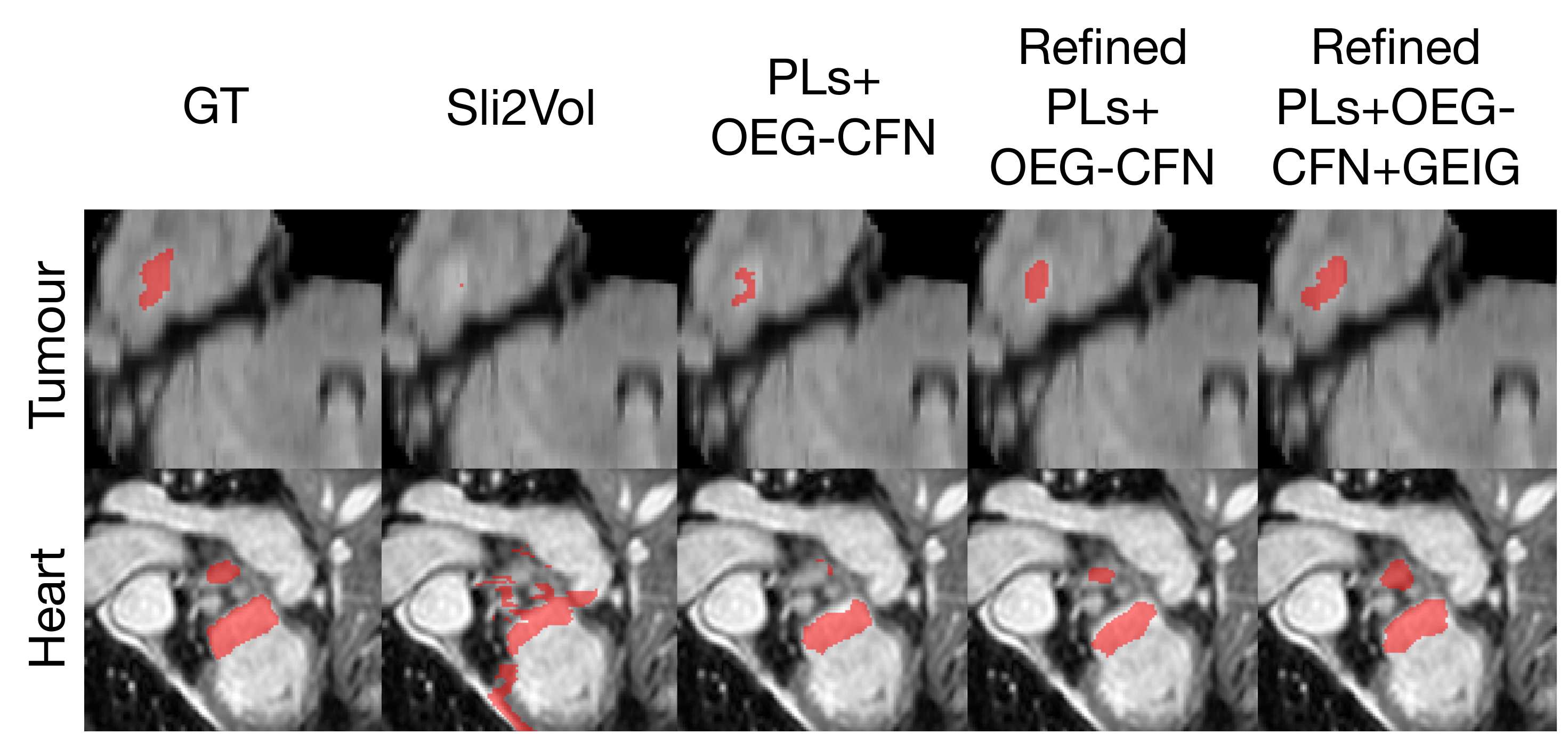}
   \caption{Examples of segmentation results generated by our method when combining different key components on the Decath-Brain Tumours and Heart datasets~\cite{simpson2019large}.}
   \label{fig:vis-abalation}
\end{figure}

\subsection{Ablation Study}
\label{subsec:abla}
To examine the effects of different key components in our method, we conduct an ablation study on both the CT and MRI datasets, as shown in Tables~\ref{table:result-ct} and~\ref{table:result-mri}. We observe the following.
(1) When using our OEG-CFN to learn the correspondences between consecutive slices and PLs, the Dice scores are improved by an average of 3.4 and 1.9 (comparing row (e) and row (g)) on the CT and MRI datasets, respectively. This demonstrates the effectiveness of including PLs to learn reliable correspondences.
(2) The performance is further improved slightly when refining the quality of the generated PLs.
(3) When applying our proposed gradient-enhanced image generator to enhance the images, the performance is further improved by an average of 1.6 and 1.2 (comparing row (h) and row (i)) on the CT and MRI datasets, respectively. This verifies the effectiveness of the gradient-enhanced image generator.

Fig.~\ref{fig:vis-abalation} presents some visual results. From these results, one can see the following.
(1) When using our OEG-CFN to learn the correspondences from the slices and PLs, the quality of segmentation results is significantly improved.
(2) When refining the PLs, the segmentation quality is further enhanced.
(3) When applying the gradient-enhanced image generator, the model is able to detect more precise edges.

\section{Conclusions}
\label{sec:concl}
In this paper, we presented a new self-supervised mask propagation framework, \proposed, {for segmenting any anatomical structures in 3D medical images using only a single annotated slice per training and testing volume.} Our \proposed can learn reliable correspondences between consecutive slices and pseudo-labels by utilizing information on estimated objects provided by PLs, effectively addressing the error drift and discontinuity issues. Experiments on nine public datasets spanning ten different SOIs demonstrated the effectiveness of our new \proposed framework.

{\small
\bibliographystyle{ieee_fullname}
\bibliography{egbib}
}

\end{document}